\documentclass[letterpaper, 10 pt, conference]{ieeeconf}
\IEEEoverridecommandlockouts
\usepackage{graphicx}
\usepackage{subcaption}
\usepackage{amsmath,amssymb}
\usepackage{hyperref}
\usepackage{xcolor}
\usepackage{comment}
\graphicspath{{figures/}}

\title{\LARGE \bf
Characterizing the onset and offset of motor imagery during passive arm movements induced by an upper-body exoskeleton}

\author{Kanishka Mitra$^{1}$, Frigyes Samuel Racz$^{2}$,  Satyam Kumar$^{1}$, Ashish D. Deshpande$^{3}$, Jos\'e del R. Mill\'an$^{1,2}$% <-this % stops a space
\thanks{$^{1}$Chandra Department of Electrical and Computer Engineering, The University of Texas at Austin, Austin, TX, USA. Email:
        {\tt\small (mitrakanishka, satyam.kumar)@utexas.edu, jose.millan@austin.utexas.edu,
        }}%
\thanks{$^{2}$Department of Neurology, The University of Texas at Austin, Austin, TX, USA. Email:
        {\tt\small fsr324@austin.utexas.edu
        }}
\thanks{$^{3}$Walker Department of Mechanical Engineering, The University of Texas at Austin, Austin, TX, USA. Email:
        {\tt\small ashish@austin.utexas.edu}}%
        }%

\begin{document}

\maketitle
\thispagestyle{empty}
\pagestyle{empty}

%%%%%%%%%%%%%%%%%%%%%%%%%%%%%%%%%%%%%%%%%%%%%%%%%%%%%%%%%%%%%%%%%%%%%%%%%%%%%%%%
\begin{abstract}

Two distinct technologies have gained attention lately due to their prospects for motor rehabilitation: robotics and brain-machine interfaces (BMIs). Harnessing their combined efforts is a largely uncharted and promising direction that has immense clinical potential. However, a significant challenge is whether motor intentions from the user can be accurately detected using non-invasive BMIs in the presence of instrumental noise and passive movements induced by the rehabilitation exoskeleton. As an alternative to the straightforward continuous control approach, this study instead aims to characterize the onset and offset of motor imagery during passive arm movements induced by an upper-body exoskeleton to allow for the natural control (initiation and termination) of functional movements. Ten participants were recruited to perform kinesthetic motor imagery (MI) of the right arm while attached to the robot, simultaneously cued with LEDs indicating the initiation and termination of a goal-oriented reaching task. Using electroencephalogram signals, we built a decoder to detect the transition between i) rest and beginning MI and ii) maintaining and ending MI. Offline decoder evaluation achieved group average onset accuracy of 60.7\% and 66.6\% for offset accuracy, revealing that the start and stop of MI could be identified while attached to the robot. Furthermore, pseudo-online evaluation could replicate this performance, forecasting reliable online exoskeleton control in the future. Our approach showed that participants could produce quality and reliable sensorimotor rhythms regardless of noise or passive arm movements induced by wearing the exoskeleton, which opens new possibilities for BMI control of assistive devices. Project page, code, and supplementary videos: \url{https://mitrakanishka.github.io/projects/passive-arm-mi/}.

\end{abstract}

%%%%%%%%%%%%%%%%%%%%%%%%%%%%%%%%%%%%%%%%%%%%%%%%%%%%%%%%%%%%%%%%%%%%%%%%%%%%%%%%
\section{INTRODUCTION}

In recent decades, two technologies have emerged with the goal of understanding and improving rehabilitation after a stroke: robotics and brain-machine interfaces (BMIs). Robot-mediated training increases afferent feedback (with high dosage and intensity), but its overall impact on neural recovery is yet low \cite{klamroth2014three}. However, if the operation of the exoskeleton was controlled by contingent neurophysiological activity, then the impact of afferent feedback would be maximized by inducing activity-dependent plasticity where neural signals and feedback are specific to each other \cite{biasiucci2018brain}. 

Recent studies combining exoskeletons and BMI have shown promising results for rehabilitation \cite{robinson2021emerging}; however,  most approaches still suffer from limited control of the robot. The control scheme often involves detecting a specific neural pattern that then triggers a pre-programmed set of movements \cite{barsotti2015full} or requires maintaining a certain state while the motion is being performed \cite{frolov2017post}. Both approaches come with their limitations: in the former, the set of executable motor actions is limited to the number of distinguishable neural patterns, while in the latter, once movement is initiated, the maintained neural patterns will be superimposed by sensory-evoked activity and instrumental noise induced by the operation of the exoskeleton, making decoding more difficult. Here we propose a novel scheme for exoskeleton operation that might allow for more precise online control and thus facilitate future applications combining BMI and robotics technology, for motor rehabilitation especially.

Motor imagery (MI) is a widely used BMI modality \cite{pfurtscheller2001motor}. During MI, a motor action is imagined without actually executing it, evoking an event-related desynchronization (ERD) and consequent decrease of spectral power in the $\mu$ (8-13 Hz) and $\beta$ (13-30 Hz) bands of their electroencephalogram (EEG). It appears as a reasonable approach that as long as the BMI user performs MI, the robot moves continuously. However, it has been shown that not only maintaining MI but also its initiation and termination have distinct EEG patterns \cite{orset2021stopping}. As argued by the authors, these phenomena could provide natural exoskeleton control: a first decoder detects MI onset as compared to the idle state for movement initiation, while a second decoder recognizes MI stop with respect to maintaining MI. The problem is that, as mentioned above, the EEG pattern associated with maintaining MI will be superimposed by other sensorimotor rhythms induced by the passive movement of the subject's limb.
In what follows, we show the feasibility of the approach despite the impact of robot motion on EEG during MI maintenance by evaluating its performance in offline and pseudo-online settings. Our results indicate that fast natural robotic exoskeleton control is possible with this hierarchical classification strategy, which, therefore, can provide a useful tool for future rehabilitation applications combining BMI and robotics.

\section{METHODS}

\subsection{Participants, Experimental Setup and Protocol}

%%%%%%%%%%%%%%%%% FIG 1: Harmony and Timeline %%%%%%
\begin{figure*}[ht!]
    \centering
    \includegraphics[width = 7 in]{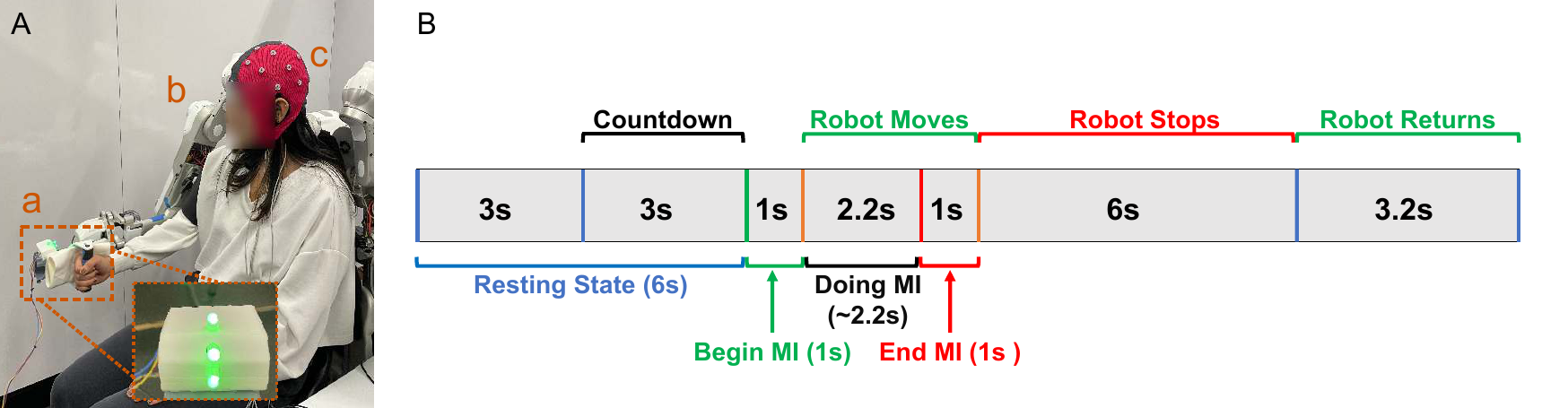}
    \caption{Experimental setup and task paradigm pipeline. Panel A shows a representative subject and the Harmony robotic exoskeleton (a. LED indicating trial phases, b. Harmony exoskeleton, c. EEG cap), while panel B shows the timeline of a task trial (see text for details).}
    \label{fig:fig1_setup_pipeline}
    \vspace{-6mm}
\end{figure*}
%%%%%%%%%%%%%%%%%%%%%%%%%%%%%%%%%%%%%%%%%%%%%%%%%%%%

Ten young, healthy individuals (age 23.90$\pm$3.78, three females, all right-handed) participated in the study. All participants provided written informed consent. The study was carried out in accordance with the Declaration of Helsinki and was approved by the Institutional Review Board at the University of Texas at Austin (2020-03-0073).

 The Harmony Exoskeleton, designed for clinical and neurological applications, was used for this study. Harmony is a bilateral rehabilitation robot with seven active degrees of freedom (DOF) per side, including five DOF for the shoulder complex. The robot can move, guide, and perturb movements of the upper limb of the participants and can record end-effector kinematics and mechanical quantities such as position, velocity, and torque \cite{kim2017upper}. In this study, Harmony performed an active goal-directed, reaching task that emphasized shoulder and elbow movement using impedance control \cite{de2019exploring}. The system also included three LEDs that were attached to Harmony's right arm (Fig. \ref{fig:fig1_setup_pipeline}A), and the colored lights would indicate the different tasks given to the participant (see below). The participant was instructed to focus on their right arm to observe the task cues for the trial but also to maintain a strong peripheral level connection \cite{biasiucci2018brain}. 

The timeline of the task paradigm is illustrated in Fig. \ref{fig:fig1_setup_pipeline}B. The starting position involved the subject sitting with their right arm by their side with an elbow flexed at a $\sim90$ degree angle while strapped to the exoskeleton. Each trial started with a 3-second resting period followed by a 3-second countdown indicated by the three white LEDs (turning on one by one every second). At the end of the countdown, all LEDs turned green, signaling the subject to begin performing MI. The specific instruction given to the subject was to imagine their right arm moving in a continuous manner (i.e., waving or reaching) without actually moving it. After a 1-second latency (1-second begin-MI period, bMI), the robot initiated the reaching movement while the subject's arm was passively extended to a horizontal reaching position. Exactly 2.2 seconds into the action, the LEDs turned red, indicating the subject to stop MI, where they were instructed to think about their arm stopping the continuous movement, and the robotic arm came to a stop after a 1-second latency (1-second end-MI period, eMI). The robotic arm remained stationary in the extended position for 6 seconds before the LEDs turned blue, and the robot returned to the starting position in 3.2 seconds, concluding one trial. The countdown period was introduced so that subjects could time the onset of MI as precisely as possible while they could learn to time MI-offset based on visual and proprioceptive feedback. Between the bMI and eMI phases, subjects were instructed to actively maintain MI (2.2-second do-MI period, dMI), while for all other periods (robot stops, returns, and resting state), they were directed to remain relaxed and not to perform any specific mental activity. Also, participants were explicitly instructed to keep their arms relaxed so that the passive movement is solely executed by the exoskeleton. This design allowed for the following comparisons: \emph{i)} onset: between resting-state (RS) and bMI, where the robot is always stationary, and \emph{ii)} offset: between dMI and eMI, where the robot is moving, and therefore, we can expect substantial induced sensorimotor activity (and plausible instrumental noise).

One measurement run consisted of 20 trials, and each participant completed 6 runs with $\sim$2-3 minutes of rest in between. A recording session lasted for about 90 minutes.

\subsection{Data Collection and Pre-processing}
EEG data were collected with an eego system (ANT Neuro, The Netherlands) at a sampling rate of 512 Hz from 22 cortical locations according to the international 10-20 system, covering mostly the sensorimotor cortex (locations F7, F3, Fz, F4, F8, FC5, FC1, FC2, FC6, C3, Cz, C4, CP5, CP1, CP2, CP6, P7, P3, Pz, P4, P8, POz). The reference electrode was placed over CPz. In addition, electrooculography (EOG) data were recorded at 512 Hz with electrodes positioned on the two temporals and above the left eye, with a common reference under the left eye. Importantly, the recording session for each participant began with collecting 90 seconds of calibration data while they were performing various eye movements, which was then used to estimate the EOG filter matrix (see below).

Signals were first band-pass filtered using a 4$^{th}$ order zero-phase Butterworth filter with cutoff frequencies of 0.1 to 45 Hz. EOG-related components (such as blinks or eye movements) were regressed out using the automated method proposed by \cite{schlogl2007fully}, where the subject-specific EOG-filter matrix was obtained from prior calibration recordings. EEG data were re-referenced to the common average reference.

\subsection{Time-Frequency Analysis and Feature Extraction}
In order to ensure that the subjects were engaged in the task, we performed a time-frequency decomposition of continuous trial data. In that, power spectral density for each channel was computed in a sliding window fashion, utilizing a window size of 0.5 seconds and a step size of $1/16$ second. In each window, the power spectrum was estimated using the Welch's method. To quantify ERD/ERS (event-related synchronization) at all frequencies on a comparable scale, raw spectral power values were transformed according to \cite{orset2021stopping} as $ERD(f)/ERS(f)=log10(A(f)/B(f))$, where $A(f)$ indicates spectral power at frequency $f$ and $B(f)$ denotes the baseline. Specifically, baseline power for each trial was obtained as the average amplitude during the one-second (resting-state) period before the countdown cue.

For motor intent classification, 1-second bMI and eMI samples were obtained as the corresponding time periods (see Figure \ref{fig:fig1_setup_pipeline}B). For the RS vs. bMI comparison, RS data was selected as the 1-second preceding the countdown cue, while for dMI data was extracted from the middle of the period when the robot was moving, and the subjects were instructed to perform MI continuously (until the eMI cue). From these 1-second-long epochs, spectral power was estimated in the same sliding window manner as described previously in the 8-30 Hz range at a one Hz resolution. This procedure yielded $22\ channels * 23\ frequencies = 506$ raw power features for each window. As each epoch provided 9 windows, this resulted in $9\ windows * 20\ trials * 6\ runs = 1,080$ samples per condition for each subject.

\subsection{Classification Approach}
Classification performances for the two tasks were assigned to the validation set, while the rest of the data was used for training the model. Subject-wise performance was assessed as the average LORO-CV classification accuracy ($\#$ of correctly classified samples/$\#$ of all samples) taken over the six CV iterations. In each run, we used only the top 10 features with  the highest Fisher scores (as obtained only from training data) for training and evaluation \cite{perdikis2018cybathlon}. Diagonal linear discriminant analysis (dLDA) was chosen as the classification method. Chance level was estimated as $54.4\%$ at significance level $\alpha=0.05$ assuming a binomial distribution of misclassifications as proposed by \cite{combrisson2015exceeding}.

\section{RESULTS}

\subsection{EEG Modulations during the Task Paradigm}

%%%%%%%%%%%%%%%%% FIG 2: Grand Spectrogram %%%%%%%%%
\begin{figure}[!t]
    \centering 
    \includegraphics[width=250pt]{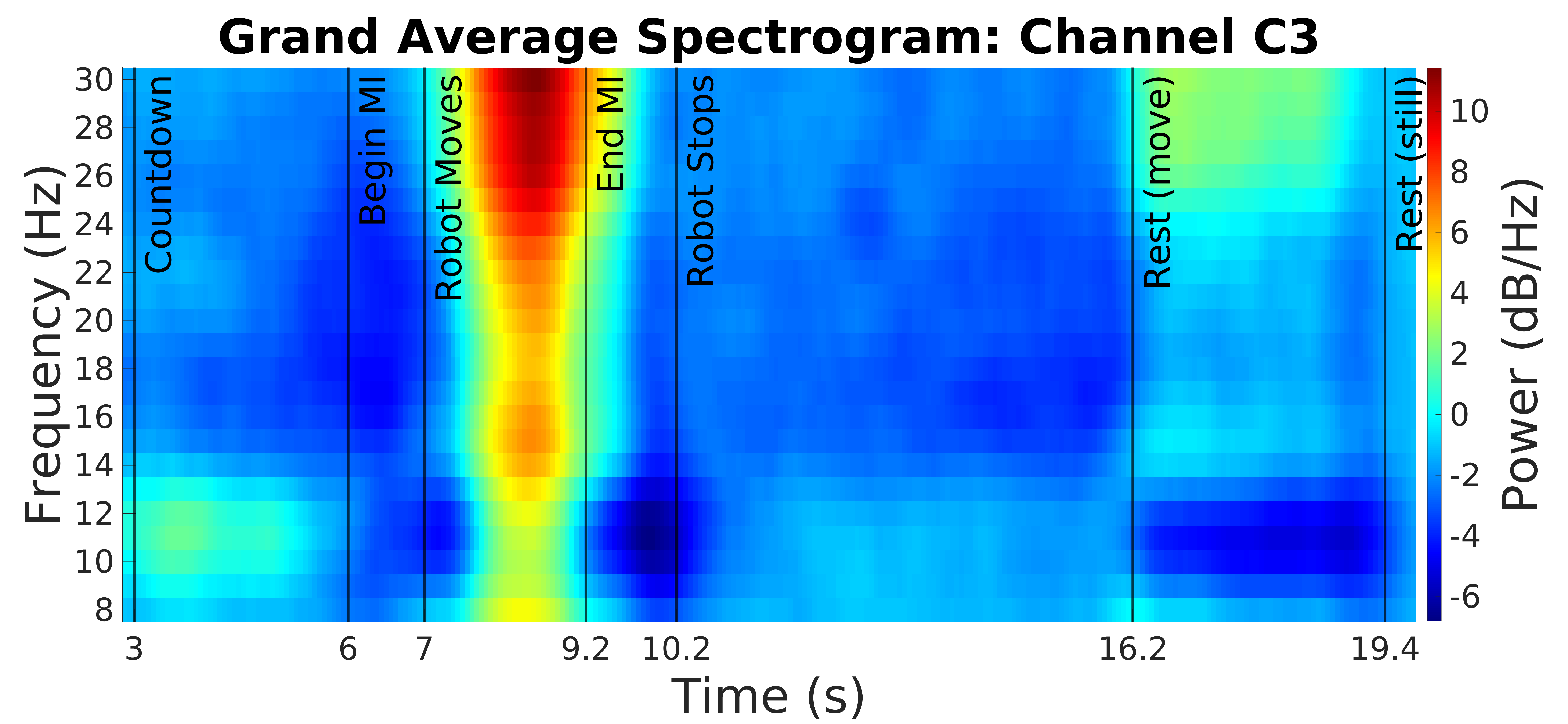}
    \caption{Grand average spectrogram from C3 region showing changes in spectral power over the course of task trials. Data was obtained by averaging over all trials and all subjects.}
    \label{fig:fig2_grand_spec}
    \vspace{-6mm}
\end{figure}
%%%%%%%%%%%%%%%%%%%%%%%%%%%%%%%%%%%%%%%%%%%%%%%%%%%%

Figure \ref{fig:fig2_grand_spec} shows how spectral power changes over the course of a trial in region C3. Compared to baseline (RS), an ERD can be observed in the $\mu$ band and in the $\beta$ band to a lesser extent during bMI, followed by an overall increase in power when the robot starts moving. This increase starts fading away once subjects engaged in eMI, with a prominent $\mu$ ERD.
In the transitory idle phase, there is practically no change in power, while a concurrent increase in high-frequency power and $\mu$ ERD can be observed while the robot returns to the start position.

\subsection{Offline Classification} 

%%%%%%%%%%%%%%%%% Table 1: Class Acc. %%%%%%%%%%%%%%%%
% results table
\begin{table}[h]
\caption{Classification accuracies}
\begin{center}\resizebox{0.8\columnwidth}{!}{  
\begin{tabular}{|c|c|c||c|c|}
\noalign{\hrule height 1.5pt}
    & \multicolumn{2}{c||}{MI-onset} & \multicolumn{2}{c}{MI-offset} \vline\\
\hline
ID & train & test & train & test\\
\noalign{\hrule height 1.5pt}
1 & $61.7\%$ & $57.0\%$ & $70.4\%$ & $64.7\%$\\
\hline
2 & $71.1\%$ & $69.3\%$ & $71.2\%$ & $68.1\%$\\
\hline
3 & $67.3\%$ & $65.9\%$ & $67.5\%$ & $63.7\%$\\
\hline
4 & $62.1\%$ & $55.7\%$ & $74.1\%$ & $64.7\%$\\
\hline 
5 & $63.0\%$ & $57.8\%$ & $70.3\%$ & $69.1\%$\\
\hline
6 & $69.7\%$ & $66.4\%$ & $71.2\%$ & $64.4\%$\\
\hline
7 & $62.4\%$ & $58.2\%$ & $70.5\%$ & $68.6\%$\\
\hline
8 & $62.1\%$ & $59.9\%$ & $71.6\%$ & $66.3\%$\\
\hline
9 & $61.9\%$ & $50.1\%$ & $69.8\%$ & $68.5\%$\\
\hline
10 & $68.6\%$ & $67.2\%$ & $69.6\%$ & $64.8\%$\\
\hline

\noalign{\hrule height 1.3pt}
\bf{AVG} & \bf{65.0\%} & \bf{60.7\%} & \bf{70.6\%} & \bf{66.6\%}\\
\hline
\bf{STD} & \bf{3.8\%} & \bf{6.2\%} & \bf{1.7\%} & \bf{2.0\%}\\
\hline
\bf{MIN} & \bf{61.7\%} & \bf{50.1\%} & \bf{67.5\%} & \bf{63.7\%}\\
\hline
\bf{MAX} & \bf{71.1\%} & \bf{69.3\%} & \bf{74.1\%} & \bf{69.1\%}\\
\noalign{\hrule height 1.5pt}
\end{tabular}}
\end{center}
\label{tb:results_LDA_table}
\end{table}
%%%%%%%%%%%%%%%%%%%%%%%%%%%%%%%%%%%%%%%%%%%%%%%%%%%

Table \ref{tb:results_LDA_table} reports the classification results of the two decoders. bMI could be distinguished from RS with better-than-chance ($>54.4\%$) accuracy for all but one subject (subject 9), with a group average accuracy of $60.7\pm6.2\%$. In comparison, eMI could be detected reliably for all subjects with an overall average accuracy of $66.6\pm2.0\%$.

%%%%%%%%%%%%%%%%% FIG 3: dLDA Acc vs. TIME %%%%%%%
\begin{figure}[!t]
    \centering
    \includegraphics[width=240pt]{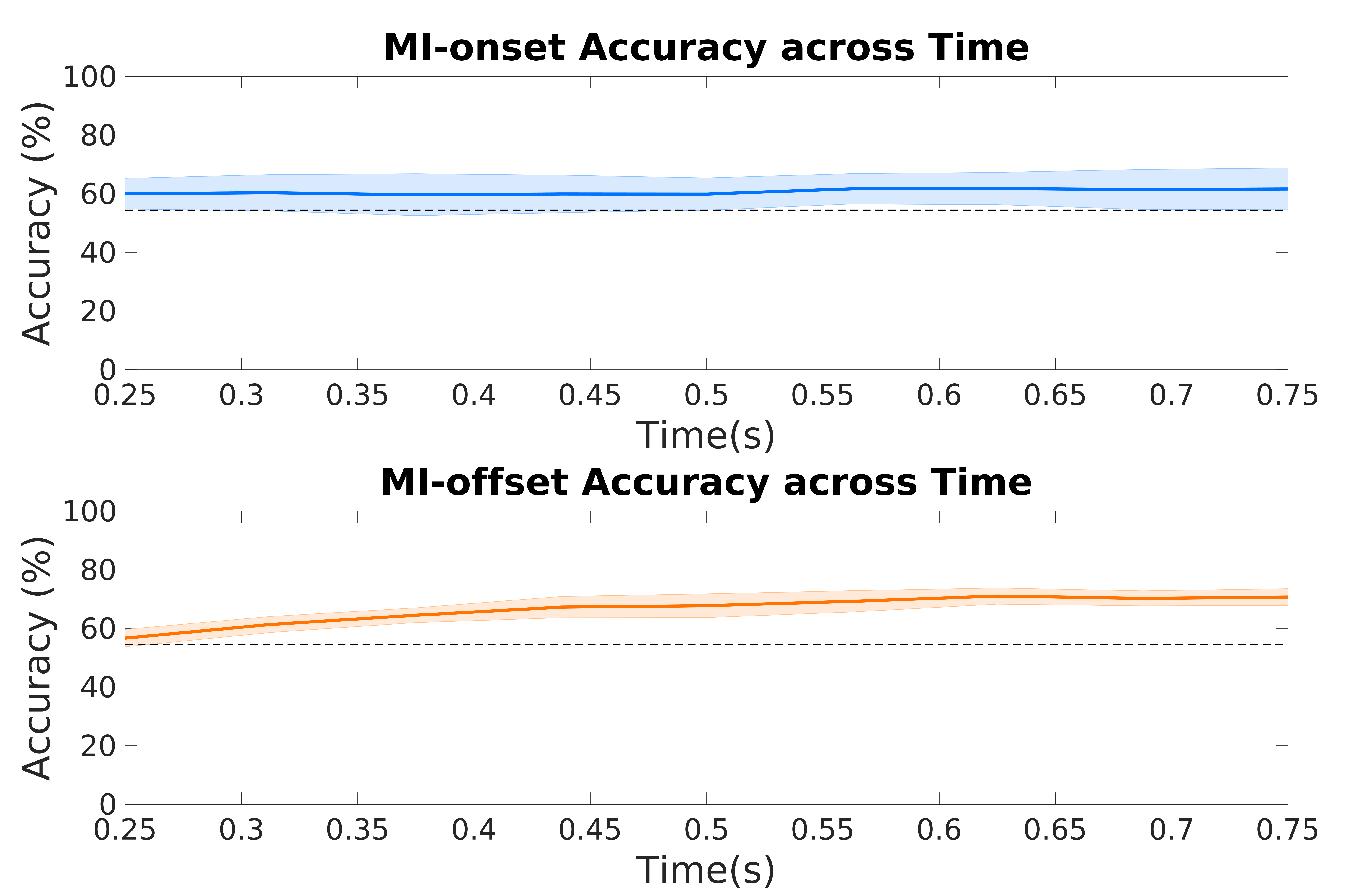}
    \caption{Grand average classification accuracies over time during the 1-second epochs. Shaded areas indicate standard deviation, and the dashed black line denotes the chance level ($54.4\%$).}
    \label{fig:fig3_accuracy_dLDA}
    \vspace{-6mm}
\end{figure}
%%%%%%%%%%%%%%%%%%%%%%%%%%%%%%%%%%%%%%%%%%%%%%%%%%%

Figure \ref{fig:fig2_grand_spec} clearly indicates that the movement of the exoskeleton induced a strong noise component. It is important to note that even though the exoskeleton was still moving during the MI-offset phase (see Fig. \ref{fig:fig1_setup_pipeline}), the movement did not finish suddenly; instead, the robot decelerated gradually until stopping. Despite this happening only very late in the 1-second frame, if the extent of contamination was proportional to movement velocity, it might have resulted in the classifier picking up on the reduced noise content in later windows. Therefore, it must be verified that classification (especially) in the MI-offset task is not based purely on the presence/absence of external confounding factors in the two conditions (dMI and eMI). To achieve that, we also investigated how classification accuracy might change over time in the 1-second windows for the two task conditions (Figure \ref{fig:fig3_accuracy_dLDA}). For MI-onset classification, performance appears to be constant over time. In contrast, there is a slight increase in accuracy over time for MI-offset detection; however, it is worth noting that performance is above chance level even for the first time window ($\sim57\%$).

\subsection{Feature Importance Analysis}

%%%%%%%%%%%%%%%%% FIG 4: Top 10 Disc. Feats %%%%%%%%
\begin{figure}[!t]
    \centering
    \includegraphics[width=240pt]{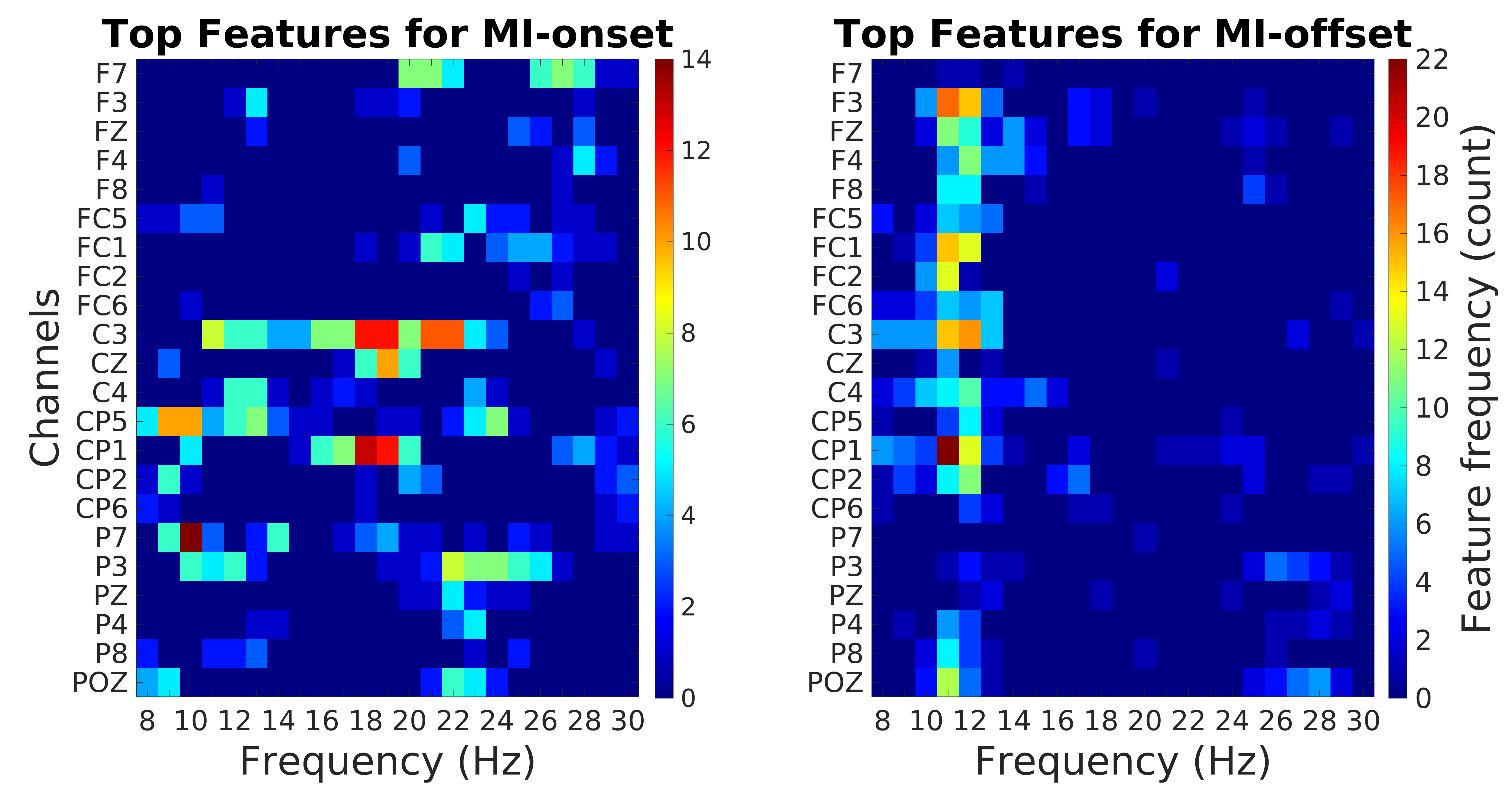}
    \caption{Top 10 most discriminative feature frequencies. The color indicates how frequently a given feature was selected as one of the top 10 for the MI-onset (left) and MI-offset (right) distinctions.}
    \label{fig:feature_importances}
    \vspace{-6mm}
\end{figure}
%%%%%%%%%%%%%%%%%%%%%%%%%%%%%%%%%%%%%%%%%%%%%%%%%%%

Finally, we wanted to explore if detection is indeed achieved based on neurophysiological features relevant to the onset and offset of MI.  Figure \ref{fig:feature_importances} shows the frequency of the top 10 most discriminate features across all the subjects for the two classification tasks. Both in the case of MI-onset (Fig. \ref{fig:feature_importances}, left) and MI-offset (Fig. \ref{fig:feature_importances}, right), the most frequently selected features were from electrodes over the somatosensory cortical regions, and in the $\mu$ frequency range corresponding to sensorimotor rhythms.

%%%%%%%%%%%%%%%%% FIG 5: Test vs. CTRL specs %%%%%%
\begin{figure*}[ht!]
    \centering
    \includegraphics[width = 7 in]{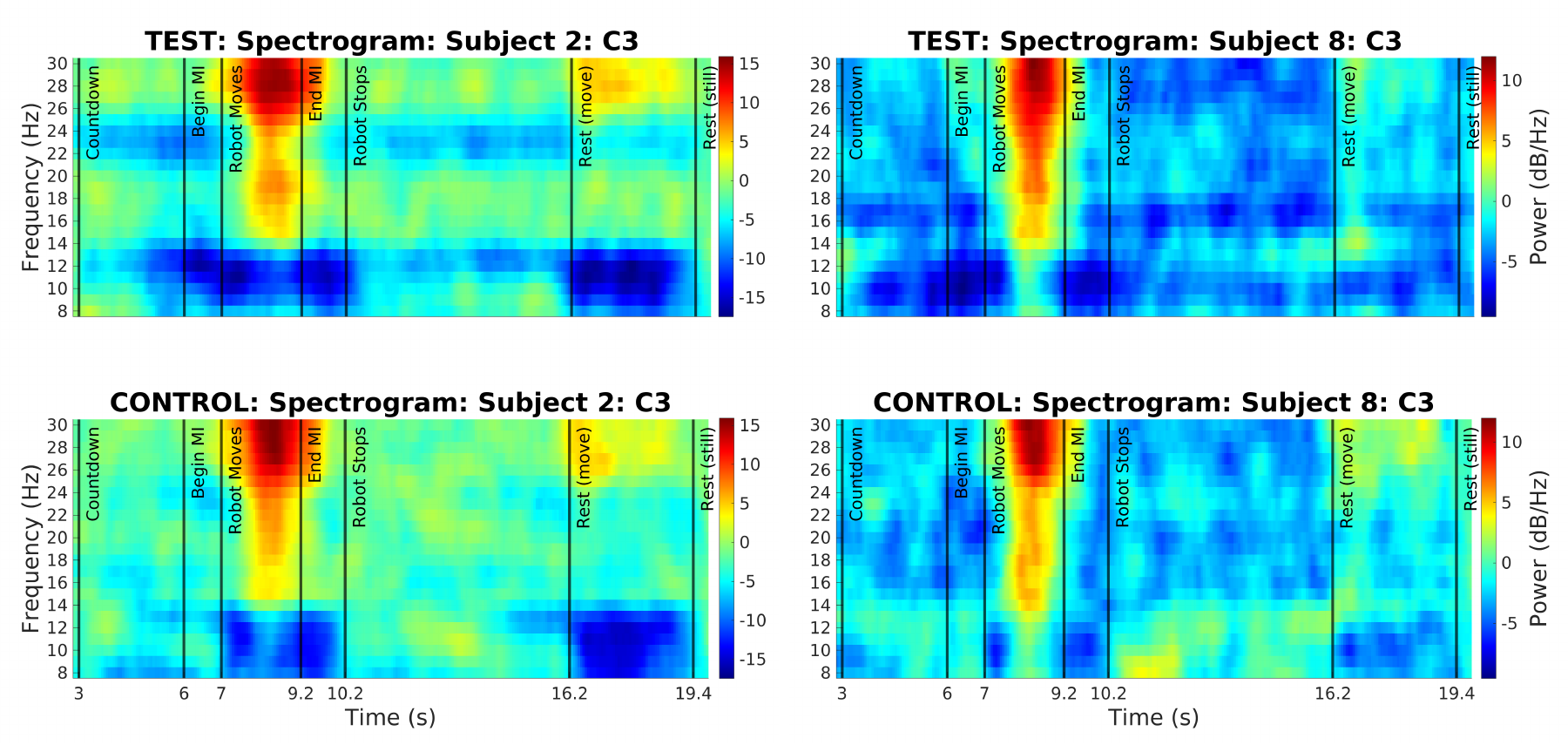}
    \caption{Average spectrograms of the two subjects (Left: subject 2; Right: subject 8) re-recorded with the task protocol that includes control trials. The top and bottom rows correspond to the test and control trials, respectively.}
    \label{fig:fig5_grand_spec_C3_test_ctrl}
    \vspace{-6mm}
\end{figure*}
%%%%%%%%%%%%%%%%%%%%%%%%%%%%%%%%%%%%%%%%%%%%%%%%%%%

\section{DISCUSSION}

In this study, we set out to detect the onset and offset of MI in individuals while the Harmony exoskeleton passively moved their arm. We identified distinct EEG patterns related to the onset and offset of MI (most prominently, a strong ERD in the $\mu$ frequency range) that we could exploit to achieve classification performances significantly better than chance level in an offline evaluation setting.

In addition to the strong $\mu$ ERD during the active period, we observed two additional patterns in Fig. \ref{fig:fig2_grand_spec}: \emph{i)} a strong, mostly broadband synchronization coinciding with the robot movement, and \emph{ii)} a similar $\mu$ ERD during the passive period when the robot returned to the home position. This raises two potential issues. First, $\mu$ ERD is associated with sensorimotor stimulation \cite{pfurtscheller1992event, pfurtscheller1999event}, which in this case is elicited by the passive arm movements driven by the exoskeleton that overlaps with MI-related activity in the active phase. Second, during the maintain-MI period, in addition to the superposition of these patterns, there seems to be external noise introduced by the moving exoskeleton.
%Furthermore, even though one might expect the same kind of artifact during the return phase (as the movement velocity is the same), the extent of the introduced noise is noticeably smaller in amplitude, as well as it does not span the entire frequency range but instead rather contaminates high-frequency activity.

Therefore, in order to confirm that subjects indeed produced MI-related activity --- and what we detect is not merely evoked sensorimotor activity or instrumental noise --- we performed additional recordings, where active MI trials were randomly interspersed with fully passive trials. In the latter, the subject was instructed not to perform MI at all and just remain relaxed until the exoskeleton completed the movement cycle. Trial timing was kept exactly the same as for active trials. We re-recruited two of the subjects of the study for these additional recordings, where they completed 6 runs comprised of 10 control (passive) and 10 test (active MI) trials in randomized order. Figure \ref{fig:fig5_grand_spec_C3_test_ctrl} illustrates average spectrograms of both conditions of the two subjects. In the test condition, $\mu$ ERD starts during the begin-MI phase, well before the exoskeleton initiates movement, indicating that subjects were actively engaging in performing MI. This $\mu$ ERD is absent in the control condition and is only elicited after movement initiation by Harmony. To test this difference more rigorously, we compared spectral power values averaged over the 1-second begin-MI period for all frequencies between the test and control trials for both subjects independently using Mann-Whitney U tests. For both subjects, spectral power was significantly lower for frequencies 9 to 13 Hz ($p<0.05$, Bonferroni-adjusted) in the active condition when compared to the control. For subject 2 (Fig. \ref{fig:fig5_grand_spec_C3_test_ctrl}, left), we also observed an MI-related ERD in the $\beta$ band (21-25 Hz) during the begin-MI phase; however, this difference did not appear significant following Bonferroni adjustment. Notably, the periods when the exoskeleton was returning to its home position exhibited identical patterns in the test and control trials. Additionally, the  instrumental noise component also appeared very similar in both conditions, further supporting that this was induced by the exoskeleton system and is not an endogenous neurophysiological activity. However, why this instrumental noise is different in the reaching and returning movements requires further investigations beyond the scope of the current study. We conjecture that it might be due to different activity in the trapezius muscle elicited by the passive arm extension and flexion that propagates to the EEG electrodes.

%%%%%%%%%%%%%%%%% FIG 6: Riemannian Accuracy %%%%%%
\begin{figure}[ht!]
    \centering
    \includegraphics[width = 240pt]{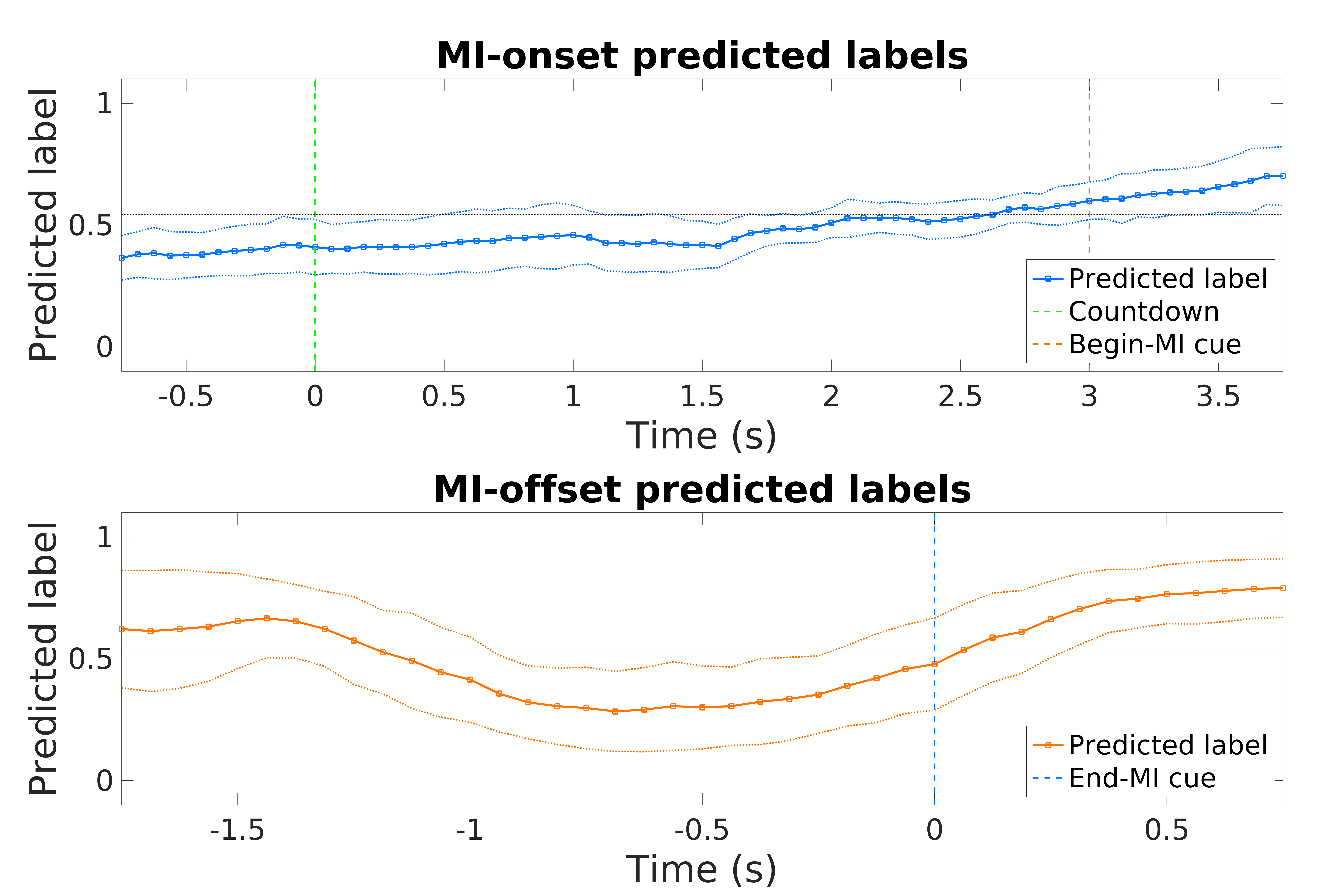}
    \caption{Time-resolved performance of Riemannian classifier. The charts show the grand average of predicted labels over time in the relevant segments of task trials. Top: RS (class 0) vs. bMI (class 1). Bottom: dMI (class 0) vs. eMI (class 1). As a reference, the horizontal dashed lines indicate chance level ($54.4\%$).}
    \label{fig:accuracy_riemannian}
    \vspace{-6mm}
\end{figure}
%%%%%%%%%%%%%%%%%%%%%%%%%%%%%%%%%%%%%%%%%%%%%%%%%%%%

The obtained results clearly indicate that a two-stage hierarchical classification approach for movement initiation and termination is viable for robotic exoskeleton control, even in the presence of instrumental noise and with the superposition of additional sensorimotor rhythms induced by passive arm movements. However, one of the main limitations of our study is that analysis was carried out offline, and our experiments did not involve online recordings where subjects had to operate the Harmony system in real time. To alleviate this issue, we performed a pseudo-online analysis of the collected data, accurately replicating an online setting in its design. For this purpose, the data was band-pass filtered using a 2$^{nd}$ order \emph{causal} Butterworth filter with cutoff frequencies from 8 to 30 Hz, and EOG activity was regressed out as-if in an online method proposed by \cite{schlogl2007fully}. Previous studies indicated that Riemannian geometry-based classifiers (RGBCs) could achieve superior performance in classifying MI-related neural patterns \cite{barachant2010riemannian, congedo2017riemannian}. In addition, RGBCs allow for online distribution matching of training and test samples, promoting robust and stable performance over multiple BMI sessions \cite{kumar2019towards}. From these considerations, we decided to use a Riemannian geometry-based, minimum distance to the mean (MDM) classifier --- which utilizes covariance matrices estimated from epoched EEG data as features --- to detect the transitions from rest to begin-MI and from maintaining-MI to end-MI in the pseudo-online pipeline. Decoders were trained on the same 1-second data segments as previously, but the performance was also evaluated continuously over the test run trials. We employed the same LORO-CV scheme; however, to replicate an online scenario, data in each validation run was processed and analyzed continuously in temporal order, applying the adaptive online re-centering method described in \cite{kumar2019towards}. Fig. \ref{fig:accuracy_riemannian} shows the results of this analysis. In general, the Riemannian approach slightly surpassed the dLDA in performance (even with less efficient pseudo-online pre-processing) with a group-average accuracy of $64.0\pm7.6\%$ and $66.6\pm7.7\%$ for MI-onset and MI-offset, respectively. Notably, performance for all subjects was above chance level. 
Fig.\ref{fig:accuracy_riemannian} shows that the RGBC successfully captures the transition from RS to bMI during the countdown period and, similarly, from dMI to eMI. Notably, for dMI, data from -1.25 to -0.25 was used for training the decoder, and indeed, in that period, the classifier mostly predicted the correct class (class 0). Surprisingly, in the preceding period --- in which the subjects are assumed to actively perform MI --- the performance deteriorates. However, this only indicates that since the classifier was trained on a different segment, it did not classify those patterns reliably as dMI. Nevertheless, these results show great potential that the hierarchical movement onset/offset approach would indeed perform well in a true online setting.

Orset and colleagues \cite{orset2021stopping} found that MI-onset was characterized with a sudden $\beta$ ERD followed by a slower, gradual $\mu$ ERD, while MI-termination exhibited an increase in $\beta$ power with $\mu$ returning to baseline. Importantly, both conditions could be reliably distinguished from RS. Here we did not observe the exact same patterns: bMI was characterized by a strong $\mu$ ERD and a weaker $\beta$ ERD, as well as we did not find a `rebound' in $\beta$ power post-eMI. These differences could be explained by the fact that the induced sensorimotor rhythms (i.e., proprioceptive feedback due to passive arm movement that gives rise to ERD) masked the $\beta$ rebound.
Thus, the EEG patterns that we found are still consistent with expectations in an MI-based paradigm.

We also found a slight increasing trend in performance toward later windows in the MI-offset detection task. This might be due to two reasons. First, even though the stop-MI cue was delivered at the same time in every trial, there was no countdown (unlike with the MI-onset case) to help subjects time their actions precisely. Therefore, we can expect some response latency after the end-MI cue delivery, which would explain why EEG activity could be better separated later in the time window.
%Second, even though the exoskeleton was still moving during the MI-offset phase, the movement did not finish suddenly; instead, the robot decelerated gradually until stopping. Even though this happened only very late in the 1-second frame, if the induced noise was proportional to movement velocity, it might have resulted in the classifier picking up on the reduced extent of noise in later windows.
Second, as already mentioned throughout the results, confounding factors introduced by the exoskeleton might be proportional to movement velocity. Unfortunately, we could not extract joint angles with the current setup to quantitatively assess this question, a limitation that we must overcome in future studies. Nevertheless, the minor nature of the increase in accuracy and the fact that classification performance was over chance level even in the first windows (Fig. \ref{fig:fig3_accuracy_dLDA}, bottom), as well as RGBC picking up on the eMI pattern right after the eMI-cue (Fig. \ref{fig:accuracy_riemannian}, bottom) suggest that this effect did not play a substantial role in separating eMI from dMI.

The biggest limitation of the current study is that it lacks a true online evaluation of our approach. Therefore, our immediate goal is to test the method in real time to confirm that subjects can indeed learn to perform BMI-driven reaching movements with the Harmony system utilizing the hierarchical classification scheme. Notably, previous research indicates that closed-loop online feedback and longitudinal training greatly facilitate subject learning and BMI skill acquisition \cite{perdikis2018cybathlon,perdikis2020brain,tonin2022learning}. Therefore, we can expect a further increase in detection performance.
This would pave the way to test our system in a clinical rehabilitation setting, focusing primarily on stroke patients. However, motor rehabilitation is relevant in other clinical conditions, such as spinal cord injury \cite{merante2020brain} or traumatic brain injury \cite{nolan2018robotic}. We anticipate that our presented method will also facilitate new interventions in a multitude of motor system pathologies.

\section{CONCLUSIONS}

Here we proposed a novel hierarchical detection scheme of MI onset and offset for signaling the initiation and termination of movement to a robotic exoskeleton. Both offline and pseudo-online analyses produced promising performances, indicating that the approach might be viable for exoskeleton control. Our method thus could provide a useful tool for future neurorehabilitation applications combining BMI technology and robotic systems.

%A conclusion section is not required. Although a conclusion may review the main points of the paper, do not replicate the abstract as the conclusion. A conclusion might elaborate on the importance of the work or suggest applications and extensions. 


%%%%%%%%%%%%%%%%%%%%%%%%%%%%%%%%%%%%%%%%%%%%%%%%%%%%%%%%%%%%%%%%%%%%%%%%%%%%%%%%

%%%%%%%%%%%%%%%%%%%%%%%%%%%%%%%%%%%%%%%%%%%%%%%%%%%%%%%%%%%%%%%%%%%%%%%%%%%%%%%%

%\section*{ACKNOWLEDGMENTS}

%JdRM and FSR acknowledge funding from the Charley Sinclair Foundation in Neurology. ADD serves as the Chief Research Officer for and has equity in Harmonic Bionics, a company that commercializes the Harmony exoskeleton.

%%%%end of acknowledgments 

\nocite{*}
% \addtolength{\textheight}{\topskip}
\bibliographystyle{ieeetr}
\bibliography{ref}

\end{document}